\documentclass[journal]{IEEEtran}
\usepackage[utf8]{inputenc}
\usepackage{subfig}

\usepackage{algorithm}
\usepackage{algpseudocode}
\usepackage{algorithmicx}
\usepackage{multirow}
\usepackage{rotating}
\usepackage{geometry}
\usepackage{xcolor}
\usepackage{amsmath,amsfonts,amssymb}

\begin{document}
\title{ACEV: Unsupervised Intersecting Manifold Segmentation using Adaptation to Angular Change of Eigenvectors in Intrinsic Dimension}
\author{Subhadip~Boral,~\IEEEmembership{Student Member,~IEEE,} Rikathi Pal,
        and~Ashish~Ghosh,~\IEEEmembership{Senior Member,~IEEE}
\thanks{Subhadip Boral is with the Department of Computer Science \& Engineering, University of Calcutta, Kolkata 700098 and ITER, Siksha O Anusandhan, Bhubaneswar 7501030, India.
E-mail: subhadipboral\_t@soa.ac.in}
\thanks{Rikathi Pal is with the A.K. Choudhury School of Information Technology, University of Calcutta, Kolkata 700106, India.
E-mail: rikathi.pal@gmail.com}
\thanks{Ashish Ghosh is with the International Institute of Information Technology Bhubaneswar, 751003, India.
E-mail: ash@isical.ac.in}}
\maketitle
\begin{abstract}
Intersecting manifold segmentation has been a focus of research, where individual manifolds, that intersect with other manifolds, are separated to discover their distinct properties. The proposed method is based on the intuition that when a manifold in $D$ dimensional space with an intrinsic dimension of $d$ intersects with another manifold, the data variance grows in more than $d$ directions. The proposed method measures local data variances and determines their vector directions. It counts the number of vectors with non-zero variance, which determines the manifold's intrinsic dimension. For detection of the intersection region, the method adapts to the changes in the angular gaps between the corresponding direction vectors of the child and parent using exponential moving averages using a tree structure construction. Accordingly, it includes those data points in the same manifold whose neighborhood is within the adaptive angular difference and eventually identifies the data points in the intersection area of manifolds. Data points whose inclusion in the neighborhood-identified data points increases their intrinsic dimensionality are removed based on data variance and distance. The proposed method performs better than $18$ SOTA manifold segmentation methods in ARI and NMI scores over $14$ real-world datasets with lesser time complexity and better stability.
\end{abstract}

\section{Introduction}
\label{sec:intro}

A manifold \cite{anowar2021conceptual} is a topological structure that is locally homeomorphic to a Euclidean space of a specific dimension. A multi-manifold structure is a complex arrangement where data points are distributed across multiple distinct manifolds rather than within a single continuous space. Manifold segmentation is a computational task focused on dividing intersecting and non-intersecting manifolds present in data into meaningful and connected segments, aiming to delineate the surface or interior of the manifold using specific criteria.
This segmentation process finds applications in diverse fields, including computer graphics, computer vision, medical imaging, and more. There are primarily two types of manifolds: intersecting and non-intersecting. Intersecting manifolds overlap with other manifolds, while non-intersecting manifolds either exist in separate regions of space or are arranged in a manner that prevents intersection.
Intersecting Manifold Segmentation holds significant value in various applications, such as medical image analysis, precise segmentation of anatomical structures, and spectral clustering \cite{5892896}. The primary job of intersecting manifold segmentation is to learn the structure of very individual manifolds and identify the region where it intersects with other manifolds. The next step is to determine which individual manifold the data points in the intersection areas belong to by comparing the structural similarity of the manifold to the data point. 
% \begin{figure*}[h]
%     \centering
%     \fbox{\includegraphics[width=\linewidth, height=2in]{Flowchart.png}}
%     \caption{Performance dependency on training phase}
%     \label{learn}
% \end{figure*}

The challenge this intersecting manifold segmentation poses is the identification of data points present in the intersection region. Now the intersecting region may be a region that involves more than two manifolds and therefore determining the number of manifolds present in the region is also a challenging task. The efficacy of two tasks depends on how individual manifold structure is learned by the algorithm because without proper learning the aforementioned tasks are hard to accomplish. The existing works handle the challenges by incorporating different approaches. The works that segment manifolds dependent on the structure of the data fail to learn the structure of the manifold. The other algorithms segment the manifolds depending on the affinity matrices where the formation of affinity matrices is limited to specific assumptions like linear structure, infinite plane, etc. The algorithms that try to understand the tangent structure of the manifold assume that other than the tangent space the data variance is zero. These assumptions fail in real-life scenarios and the performance of the algorithms drops. This encourages the proposed method to segment intersecting manifolds with better efficiency, lesser complexity, and stability.
The proposed work, ACEV, is a two-step algorithm, where in the first step the non-intersecting manifolds are segmented and in the next step, individual intersecting manifolds are segmented. ACEV incorporates eigenvalue decomposition of the Laplacian graph matrix for the segmentation of non-intersecting manifolds. ACEV learns the intrinsic dimension of individual manifolds along with their structure and gradual structural changes. The algorithm identifies intersecting areas in a manifold by understanding the abrupt structural change in the neighborhood while traversing the data points and maintaining a tree structure. Then depending on the structural property it determines in which manifold the data points from the intersecting area lie.

The proposed method is an unsupervised manifold segmentation mechanism that initially finds the non-intersecting manifolds present in a dataset. Now, for each non-intersecting components or manifolds it segments the present intersecting manifolds by studying their intrinsic dimension. By learning the intrinsic dimension, it identifies the data points which lie in intersecting regions. Depending on eigenvalue it filters those data points and includes the data points lying in the exact manifold. The performance and effectiveness of the ACEV eclipses other existing algorithms over $11$ datasets. The performance analysis shows the independence of the proposed method over manual parameters. The following sections discuss the state-of-the-art algorithms in this research field, discuss the detailed proposed method, and give a proper performance and comparative analysis.
The following sections of the paper include discussions on state-of-the-art algorithms in the research field, provide a detailed explanation of the proposed method, and present a comprehensive performance analysis for comparison.

\section{Related Works}
The problem of intersecting manifold segmentation is addressed by various methods, which broadly belong to these categories. 
\subsection{Traditional Methods}
Methods that do not consider the structure of the individual manifolds and instead of that, segment depending on the local structure of the data. A comparative detail of few these methods are presented in Table \ref{R1}.
\begin{table*}[] 
\caption{Comparative discussion on clustering methods for manifold segmentation}
\label{R1}
\resizebox{\linewidth}{!}{%
\begin{tabular}{|c|c|c|c|c|}
\hline
\textbf{Algorithm} & \textbf{Year} & \textbf{Methodology} & \textbf{\begin{tabular}[c]{@{}c@{}}Dedicated Intersecting \\ Manifold Segmentation\end{tabular}} & \textbf{Drawbacks Addressed by ACEV} \\ \hline
$k$means \cite{jain1988algorithms} & 1988 & \begin{tabular}[c]{@{}c@{}}Clusters datapoints into $k$ clusters by minimizing the\\ distance between datapoints and cluster center.\end{tabular} & No & \begin{tabular}[c]{@{}c@{}}Apriori knowledge of \\number of manifolds\end{tabular} \\ \hline
DBSCAN \cite{ester1996density} & 1996 & \begin{tabular}[c]{@{}c@{}}Groups closely packed points based on a specified distance\\ threshold and a minimum number of points within\\ the neighborhood.\end{tabular} & No & \begin{tabular}[c]{@{}c@{}}Clustering of data with\\ non-uniform density\end{tabular} \\ \hline
BIRCH \cite{zhang1997birch} & 1997 & \begin{tabular}[c]{@{}c@{}}Performs hierarchical clustering and reduces\\ the data dimensionality.\end{tabular} & No & \begin{tabular}[c]{@{}c@{}}Apriori knowledge\\ of number of manifolds\end{tabular} \\ \hline
OPTICS \cite{ankerst1999optics} & 1999 & \begin{tabular}[c]{@{}c@{}}Generates a reachability plot to reveal the\\ hierarchical structure of the data.\end{tabular} & No & \begin{tabular}[c]{@{}c@{}}Apriori knowledge of \\number of manifolds\end{tabular} \\ \hline
Mean shift \cite{derpanis2005mean} & 2005 & \begin{tabular}[c]{@{}c@{}}Clusters by iteratively shifting data points toward\\ the mode of their local density distributions.\end{tabular} & No & \begin{tabular}[c]{@{}c@{}}Apriori knowledge of \\number of manifolds\end{tabular} \\ \hline
Spectral clustering \cite{5892896} & 2011 & \begin{tabular}[c]{@{}c@{}}Clusters data points based on the eigenvectors of\\ the affinity matrix, which is based on Euclidean\\ distances between data points.\end{tabular} & No & \begin{tabular}[c]{@{}c@{}}Segmentation of\\ intersecting structures\end{tabular} \\ \hline
\end{tabular}%
}

\end{table*}

\subsection{Manifold Structure Learning Methods}

Various methods learn the structure of manifolds and they are used for manifold segmentation. Table \ref{R2} holds the details of these methods.
\begin{table*}[] 
\caption{Comparative discussion on manifold structure learning methods for manifold segmentation}
\label{R2}
\resizebox{\linewidth}{!}{%
\begin{tabular}{|c|c|c|c|c|}
\hline
\textbf{Algorithm} & \textbf{Year} & \textbf{Methodology} & \textbf{\begin{tabular}[c]{@{}c@{}}Dedicated Intersecting \\ Manifold Segmentation\end{tabular}} & \textbf{Drawbacks addressed by ACEV} \\ \hline
Principal component analysis (PCA) \cite{wold1987principal} & 1987 & \begin{tabular}[c]{@{}c@{}}Embeds data points while preserving \\ structural information through data variance.\end{tabular} & No & \begin{tabular}[c]{@{}c@{}}Assumption of linear \\ manifold structure\end{tabular} \\ \hline
Locally Linear Embedding (LLE) \cite{roweis2000nonlinear} & 2000 & \begin{tabular}[c]{@{}c@{}}Preserves neighborhood-based \\ reconstruction weightages during embedding.\end{tabular} & No & \begin{tabular}[c]{@{}c@{}}Non-uniform manifold \\ structure embedding\end{tabular} \\ \hline
Multi Manifold Discriminant (MMD Isomap) \cite{yang2016multi} & 2016 & \begin{tabular}[c]{@{}c@{}}Uses class information for manifold \\ segmentation and manifold structure learning.\\ Also embeds each manifold using Isomap.\end{tabular} & Yes & \begin{tabular}[c]{@{}c@{}}Class information based\\ manifold segmentation\end{tabular}\\ \hline
Semi-supervised Multi Manifold Isomap (SSMM Isomap) \cite{zhang2018semi} & 2018 & \begin{tabular}[c]{@{}c@{}}Learns local linear structure and embeds data \\ points by minimizing intraclass distance and \\ maximizing interclass distance while using \\ partially labeled data.\end{tabular} & Yes & \begin{tabular}[c]{@{}c@{}}Uses class information to\\ understand manifold structure.\end{tabular} \\ \hline
UMD Isomap \cite{gao2022unsupervised} & 2022 & \begin{tabular}[c]{@{}c@{}}Captures the non-linear global relationship \\ between data points using Mixture of Probabilistic \\ PCA \cite{tipping1999mixtures}. Also determines local \\ tangent subspaces to understand the local geometry \\ of the submanifolds.\end{tabular} & Yes & \begin{tabular}[c]{@{}c@{}c@{}}Non-adaptive to the structural \\changes occurred  due to the\\ non linearity of the manifold\end{tabular} \\ \hline
\end{tabular}%
}

\end{table*}

\subsection{Individual Manifold Structure Learning Based Methods}

There are methods dedicated to segment intersecting manifolds with various assumptions and constraints. The brief discussion of these methods are given in Table \ref{R3}.

\begin{table*}[]
\caption{Comparative discussion on individual manifold structure learning methods}
\label{R3}
\resizebox{\textwidth}{!}{%
\begin{tabular}{|c|c|c|c|c|}
\hline
\textbf{Algorithm} & \textbf{Year} & \textbf{Methodology} & \textbf{\begin{tabular}[c]{@{}c@{}}Dedicated Intersecting \\ Manifold Segmentation\end{tabular}}& \textbf{Drawbacks addressed by ACEV} \\ \hline
$k$-plane clustering ($k$PC) \cite{bradley2000k} & 2000 & \begin{tabular}[c]{@{}c@{}}Clusters are obtained using hyperplanes \\ constructed by eigenvalue decomposition \\ of the affinity matrix.\end{tabular} & No & \begin{tabular}[c]{@{}c@{}}Assumptions of linear \\ separability of manifolds\end{tabular} \\ \hline
$k$FC \cite{tseng2000nearest} & 2000 & \begin{tabular}[c]{@{}c@{}}Finds $q$ linear planes which have\\ the minimum Euclidean distance from the datapoints \end{tabular}& No & Linear representation of the global structure \\ \hline
Spectral multi-manifold Clustering (SMMC) \cite{wang2011spectral} & 2011 & \begin{tabular}[c]{@{}c@{}}Measures cohesion within manifolds and \\ separability between manifolds through \\ tangent space similarity and dissimilarity.\end{tabular} & Yes & \begin{tabular}[c]{@{}c@{}}Fragile manifold structure in the intersection \\ area due to high affinity with other manifolds.\end{tabular} \\ \hline
Localized $k$-flat clustering (L$k$FC) \cite{wang2011localized} & 2011 & \begin{tabular}[c]{@{}c@{}}Introduces localized representations of \\ linear models and a distortion measure \\ for cluster quality assessment.\end{tabular} & No & \begin{tabular}[c]{@{}c@{}}Prior assumptions made \\ on linear manifold structure.\end{tabular} \\ \hline
$k$-proximal plane clustering ($k$PPC) \cite{shao2013proximal} & 2013 & \begin{tabular}[c]{@{}c@{}}Finds the best-fit orientation for planar \\ representation of the manifold by maximizing \\ the margin between manifolds.\end{tabular} & No & \begin{tabular}[c]{@{}c@{}}Linear structural representations \\ of non-linear manifolds.\end{tabular} \\ \hline
Local $k$-proximal plane clustering (L$k$PPC) \cite{yang2015local} & 2015 & \begin{tabular}[c]{@{}c@{}}Laplacian graph matrix-based connectivity \\ determination of manifolds.\end{tabular} & No & \begin{tabular}[c]{@{}c@{}}Senses intersecting \\ manifolds as a single manifold.\end{tabular} \\ \hline
Twin support vector machine for clustering (TWSVC) \cite{wang2015twin} & 2015 & \begin{tabular}[c]{@{}c@{}}Segments planar representations of \\ K-means derived clusters by incorporating \\ L2 Norm distances.\end{tabular} & No & \begin{tabular}[c]{@{}c@{}}Apriori knowledge \\ of number of manifolds.\end{tabular} \\ \hline
TVG$+$Ambiguity Resolution (TVG+AR) \cite{deutsch2015intersecting} & 2015 & \begin{tabular}[c]{@{}c@{}}Zero eigenvalue-based determination of tangents \\ and learning of data variance gap through the \\ exponential moving average helps to segment \\ intersecting manifolds.\end{tabular} & Yes & \begin{tabular}[c]{@{}c@{}}Prior knowledge of intrinsic \\ dimension of the manifold \\ and inefficiency with higher \\ dimensional data.\end{tabular} \\ \hline
\begin{tabular}[c]{@{}c@{}}L1-norm distance minimization \\ based robust TWSVC (RTWSVC) \cite{ye2017l1} \end{tabular}& 2017 & \begin{tabular}[c]{@{}c@{}}Improvement of TWSVC by incorporating \\ L1 norm distances.\end{tabular} & No & \begin{tabular}[c]{@{}c@{}}Apriori knowledge \\ of number of manifolds.\end{tabular} \\ \hline
$k$-subspace discriminant clustering ($k$SDC) \cite{li2019robust} & 2019 & \begin{tabular}[c]{@{}c@{}}Partition-based clustering of subspaces \\ and learning of local structure through L1 Norm.\end{tabular} & No & \begin{tabular}[c]{@{}c@{}}Segments region of interests instead \\of manifold structure learning\end{tabular} \\ \hline
Multiple flat projections clustering (MFPC) \cite{bai2021multiple} & 2021 & \begin{tabular}[c]{@{}c@{}}Local linear projection-based global structure \\ learning and intersecting cluster determination \\ through non-convex optimization.\end{tabular} & Yes & \begin{tabular}[c]{@{}c@{}}Inefficiency while addressing \\ non-uniform manifold structures.\end{tabular} \\ \hline
Graph MMC \cite{trillos2023large} & 2023 & \begin{tabular}[c]{@{}c@{}}Uses the Graph Laplacian operator to ensure \\intramanifold connectivity and intermanifold sparcity \end{tabular}  & Yes & \begin{tabular}[c]{@{}c@{}}Aprior knowledge and assumptions \\ about the intersecting manifolds \end{tabular}\\ \hline
% \begin{tabular}[c]{@{}c@{}}Robust local K-proximal plane clustering\\ based on L2,1-norm minimization (RLKPPC) \cite{wang2024robust}\end{tabular}  & 2024 & - & - & - \\ \hline
\end{tabular}%
}
\end{table*}

The existing works are dependent on the user parameters and fail to address major challenges of the field which include sensitivity to data intersections, parameter tuning, and computational complexity. The existing works perform mostly in $O(n^4)$ and $O(n^5)$ which motivates ACEV to perform effectively and efficiently.

%\subsection{Traditional Methods}

%\subsection{Manifold Structure Learning Based Methods}

%\subsection{Clustering Based Manifold Segmentation Methods}

\section{Proposed Work}
\subsection{Problem Statement}

Suppose there are $n$ data points represented in $D$ dimensional space and these $n$ data points are lying on $m(\geq 1)$ non-intersecting manifolds, where $l_i$ represents $i^{th}$ non-intersecting manifold. Each $l_i$ consists of $q_i(\geq 1)$ intersecting manifolds and $M_{ij}$, the $j^{th}$ individual manifold in $l_i$ has intrinsic dimension $d_{ij}$ and $1\leq i\leq m, 1\leq j\leq q_i$ and $1\leq d_{ij}\leq D$. The proposed method first segments those $m$ non-intersecting manifolds and in the second step, it separates each individual manifold $M_{ij}$, which intersects with other manifolds in $l_i$ if $q_i$ $>$ $1$, based on an unsupervised approach. This manuscript proposes an innovative approach to this.

% \begin{tabular}{|c|c|}
% \hline
% Abbreviations & Meaning \\
% \hline
% $N$ & Number of non-intersecting manifolds \\
% $m_1,m_2,m_3,m_4....m_N$ & Each non-intersecting manifolds \\
% $S$ & Parent coordinate status \\
% $d$ & Dimension of parent coordinate \\
% \hline
% \end{tabular}

\subsection{ Segmentation of non-intersecting manifolds}

The segmentation of non-intersecting manifolds is based on the work \cite{boral2021unsupervised}, which is an unsupervised manifold segmentation mechanism. The method uses a graph-based component analysis to determine the number of components or non-intersecting manifolds present in the data and uses agglomerative clustering to group data points that belong to the same manifold.

Initially, a $k-$neighborhood is found for every data point and those $k$ data points are considered adjacent to that data point. Following this mechanism an adjacency matrix is created which resembles a graph. There are $n$ data points and therefore an $n\times n$ adjacency matrix $G$ will be obtained. The method follows the idea that the singular value decomposition of a Laplacian graph matrix will depict the number of disjoint components present in the graph. In other words, the number of eigenvectors with corresponding zero eigenvalues of the Laplacian graph matrix will be the number of components in the graph. Therefore, the number of components present in the graph is found by constructing the adjacency matrix $G$ and corresponding Laplacian graph matrix $L_{G}$. In other words, as there are $m$ non-intersecting manifolds present in the dataset, the number of components in the graph $G$ will be also $m$ and the number of eigenvectors with corresponding zero eigenvalues of the Laplacian graph matrix will be $m$. The Laplacian graph matrix $L_{G}$  corresponding to the adjacency matrix $G$  will be a $n \times n$ matrix where value of the $j$th column of the $i$th row will be 
% \begin{equation}
% \label{matrix}
% \chi_{ij} =
%     \begin{cases}
%         -1  & \text{v}_{i} \text{ and {v}_{j} \text{ are adjacent}\\
%         d(v_i) & \text{i = j} \\
%         0 & \text{otherwise} 
%     \end{cases}
% \end{equation}

\begin{equation}
\label{matrix}
\chi_{ij} =
    \begin{cases}
        -1 & \text{if $v_i$ and $v_j$ are adjacent}\\
        d(v_i) & \text{if $i = j$} \\
        0 & \text{otherwise} ;
    \end{cases}
\end{equation}

where $v_i$ and $v_j$ are the $i$-th and $j$-th vertex and d($v_i$) is degree of $v_i$.
Then singular value decomposition (SVD) of the Laplacian graph matrix $L_{G}$ is performed and the number of zero eigenvalues is counted. 

Now to find which datapoint is part of which non-intersecting manifold, hierarchical agglomerative clustering \cite{murtagh2012algorithms} is performed on the data. This yields $m$ clusters, where each cluster represents individual non-intersecting manifolds $l_i$ and the proposed method finds the intersecting manifolds present in each $l_i$.

\subsection{Segmentation of intersecting manifolds}

\begin{figure} 
    \centering
  \subfloat[Plane intersecting with Plane\label{11a}]{%
       \includegraphics[height = 1.7in, width=0.5\linewidth]{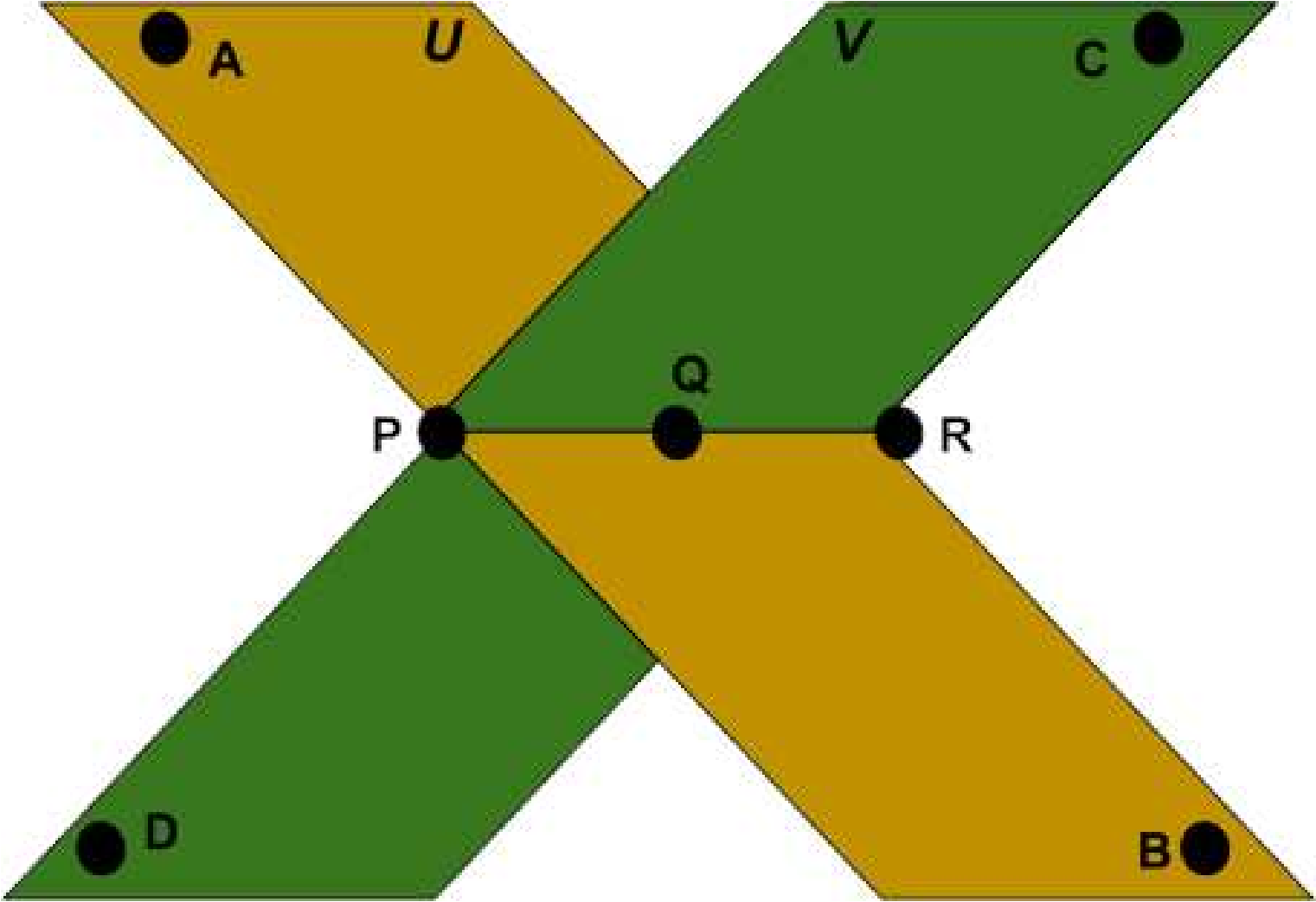}}
    \hfill
  \subfloat[Plane intersecting with S Curve\label{11b}]{%
        \includegraphics[height = 1.7in, width=0.5\linewidth]{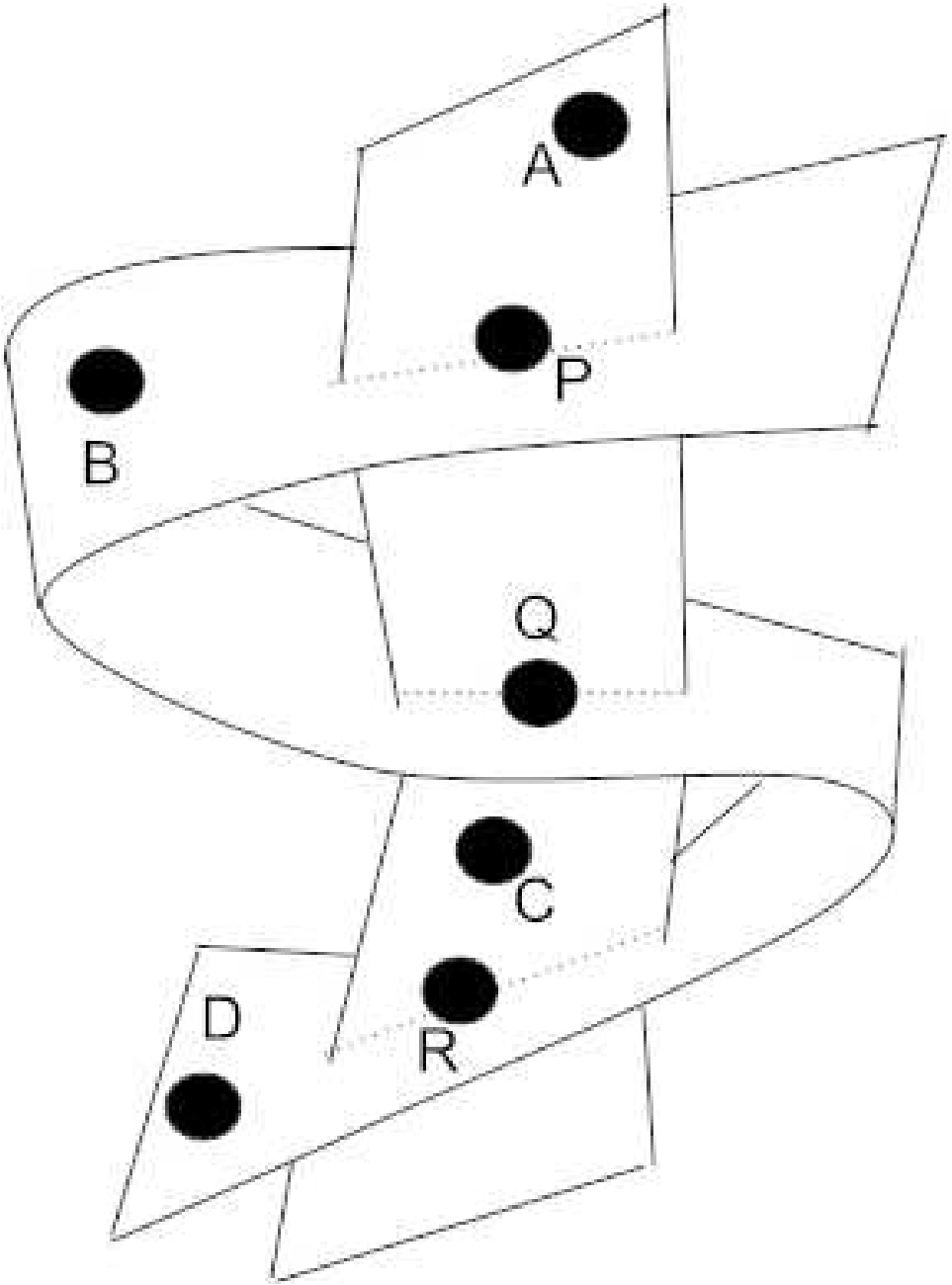}}
  \caption{Synthetic intersecting manifold structure}
  \label{Fig1} 
\end{figure}

%\subsubsection{Outline: }
The proposed method introduces a novel unsupervised intersecting manifold segmentation mechanism. In Figures \ref{11a} and \ref{11b}, data points $A$ and $B$ are attributed to manifold $U$, data points $C$ and $D$ to manifold $V$, while data points $P, Q$ and $R$ reside within the intersection of these two manifolds. The primary objective is to identify the data points situated in the intersection region of different manifolds with an approach that relies on the intrinsic dimension \cite{CAMASTRA201626} of individual manifolds. As depicted in Figures \ref{11a} and \ref{11b}, manifolds $U$ and $V$ have an intrinsic dimension $2$ and this is true for data points $A, B, C$, and $D$ as their local neighbourhood properties can be accounted in $2$-D space without information loss. However, data points $P, Q$ and $R$ possess an intrinsic dimension of $3$ as these lie in the intersection region and local neighbourhood properties can not be accounted in less than $3$-D space without information loss.

This signifies that the algorithm must learn the intrinsic dimension of individual manifold $M_{ij}$. Once learned, the algorithm can identify the data points that belong to the intersection region by comparing the intrinsic dimension. If the algorithm determines the intrinsic dimension of the manifold as $d$, data points in the intersection region will have an intrinsic dimension $>d$ as there will be more information that can not be accounted for in $d$-dimensional space due to the inclusion of data points from another manifold in their neighbourhood. The following learning process outlines the steps required to accomplish this task.

% \begin{figure}
% \includegraphics[width=\columnwidth,height = 1.5in]{1.eps} 
% \caption{Synthetic intersecting manifold structure}
% \label{Fig1}
% \end{figure}
% \begin{figure}
% \includegraphics[width=\columnwidth, height=2.5in]{alpha.eps} 
% \caption{Synthetic intersecting manifold structure}
% \label{Fig11}
% \end{figure}
% \begin{figure*}[t]
%     \centering\fbox{
%     \begin{minipage}{0.6\textwidth}
%         \centering
%         {\includegraphics[width=\linewidth]{1.eps}}
%     \end{minipage}%
%     \begin{minipage}{0.4\textwidth}
%         \caption{Synthetic intersecting manifold structure}
%         \label{Fig1}
%     \end{minipage}
%     }
% \end{figure*}

\subsubsection{Intrinsic Dimension Determination}

Suppose there is a $d$-dimensional manifold in a $D (D > d)$-dimensional vector space and the neighbourhood of a data point $t$ from that manifold is considered. Now, the covariance matrix of that neighbourhood is computed and $D$ eigenvalues and associated $D$ eigenvectors are found. The number of eigenvectors corresponding to non-zero eigenvalues for that neighbourhood will be $d$. This is true because there will be zero data variance in the other $(D-d)$ directions, which means that the associated eigenvalues will be zero. Further in the discussion, eigenvector and eigenvalues, eigenvector and eigenvalues of the datapoint and eigenvector and eigenvalue of the neighbourhood will mean eigenvectors and eigenvalues of the covariance matrix of that data point's neighbourhood. The algorithm determines the directions of the eigenvectors with non-zero eigenvalues for the neighbourhood of $t$.
Suppose, two data points $t$ and $f$ belong to the same manifold, then their neighbourhood structure will be similar and the direction of data variance will be similar and therefore, their corresponding directions of eigenvectors will be similar, i.e., $g^{th}$ principal component of datapoints $t$ and $f$ will be in the same direction. Therefore, the angular gap between two corresponding eigenvectors derived from the neighbourhood of two individual data points is decisive in determining whether those two data points belong to the same local structure or not. So, the angular gaps between the corresponding eigenvectors of the data points are calculated using equation \ref{ang12} for the data point $t$ and $f$, where $\vec{p}_{tg}$ and $\vec{p}_{fg}$ are the $g^{th}$ principal component of $t^{th}$ and $f^{th}$ data point respectively.

\begin{equation}
\label{ang12}\begin{split}
  {angle\_differ}_{g}(t, f) = \cos^{-1}\left(\frac{\langle\vec{p}_{tg}, \vec{p}_{fg}\rangle}{\|\vec{p}_{tg}\| \cdot \|\vec{p}_{fg}\|}\right),\\
  g=1, \cdots, D
  \end{split}
\end{equation}

\begin{figure}[h]  % Correct way to specify figure placement
    \centering  % To center the figure
    \includegraphics[width=0.9\linewidth]{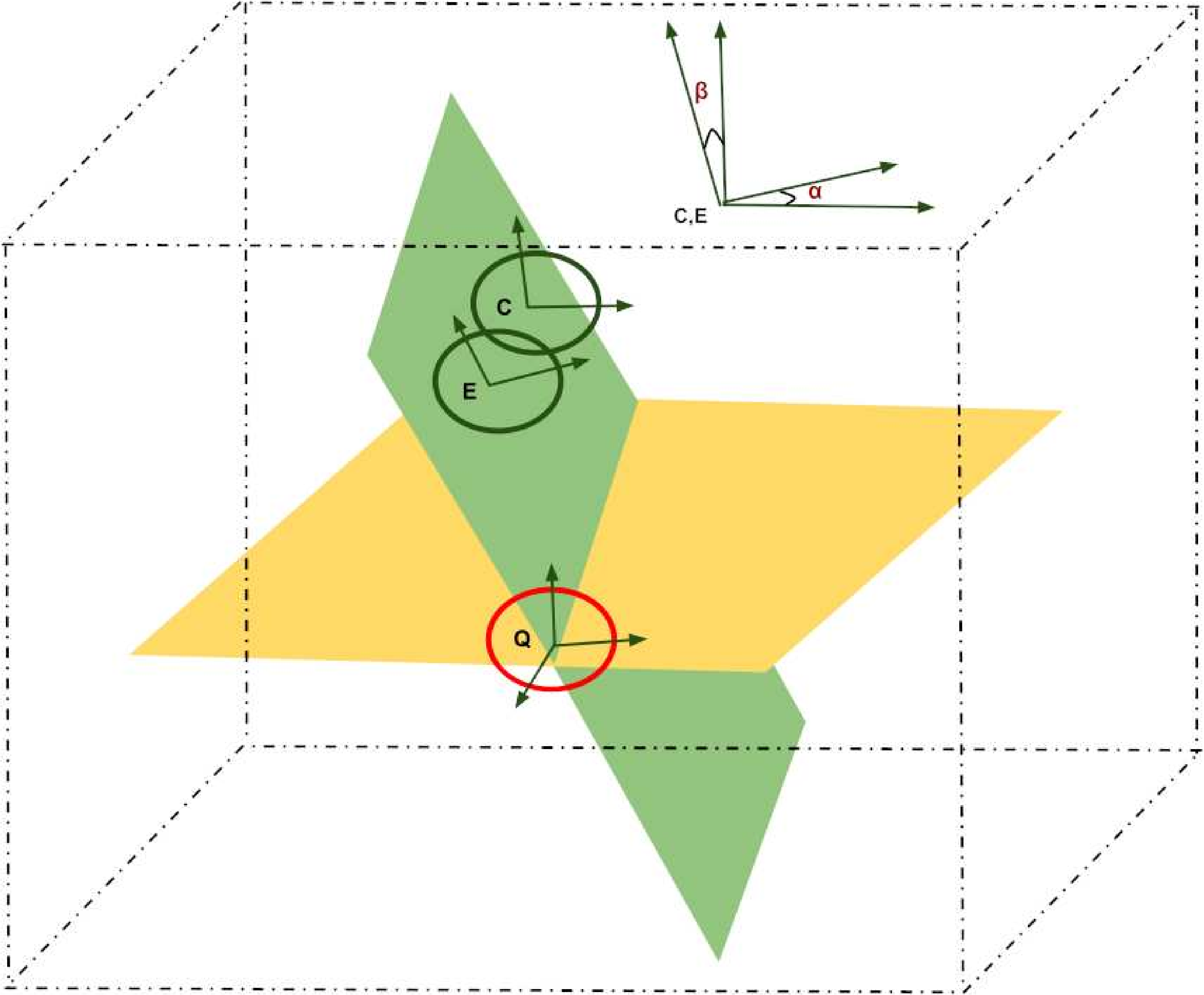}  % Correct specification of the image width
    \caption{Neighborhood (using circles) and direction of data variance (using arrows) representation of individual data points}
    \label{Fig 2}
\end{figure}

In Figure \ref{Fig 2}, the green circles show the neighbourhoods of data points $E$ and $C$, while the red circle corresponds to the neighbourhood of datapoint $Q$. The arrows within each circle denote the direction of non-zero data variance for their respective neighbourhoods, where the arrow length is independent of the amount of data variation. The angular difference between the $1^{st}$ principal components of datapoints $E$ and $C$ is denoted by $\alpha$ and the same is denoted by $\beta$ for the $2^{nd}$ principal components as shown in Figure \ref{Fig 2}. For datapoints $E$ and $C$, $\alpha$ and $\beta$ are observed to be nearly zero as the neighbourhood structures are similar. However, for datapoint $Q$, the angular difference in the third direction is high, as $Q$ is from a region which has an intrinsic dimension of $3$. It will have a non-similar $3$rd eigenvector direction when compared with the corresponding eigenvector direction of $E$ or $C$. This gap indicates a distinct structural dissimilarity compared to the other two datapoints and this signifies the change in intrinsic dimensionality.

% \begin{figure}[h]
%          {\includegraphics[width=\linewidth, height=2in]{3.eps}}
%         %\caption{Caption for Image 2}
        
%         \caption{Neighborhood (using circles) and direction of data variance (using arrows) representation of the synthetic intersecting manifolds}
%         \label{Fig 2}
% \end{figure}

% \begin{figure*}[t]
%     \centering
%     \fbox{
%     \begin{minipage}{0.6\textwidth}
%         \centering
%         {\includegraphics[width=\linewidth]{3.eps}}
%     \end{minipage}%
%     \begin{minipage}{0.4\textwidth}
%         \caption{Neighborhood (using circles) and direction of data variance (using arrows) representation of the synthetic intersecting manifolds}
%         \label{Fig 2}
%     \end{minipage}
%     }
% \end{figure*}

The mechanism mentioned above captures the intrinsic dimension of local structure, i.e., neighbourhood of data points. This also emphasizes the fact that with non-significant ${angle\_differ}_{g}(t, f), \forall g$ ACEV should include $f$ in the same manifold as $t$. However, a manifold exhibits a local resemblance to Euclidean space while forming a non-linear structure globally. Therefore, the local structural differences may be non-significant but not uniform across the entire space. Therefore, the proposed method must learn the change in the local structure in terms of angular gap and this local structural change is caused by the global non-linearity of the manifold. The algorithm learns the changes in the angular gap between eigenvectors to address this problem.

\subsubsection{Manifold Structure Learning using Time Series Analysis}
These local structural changes manifest gradually rather than abruptly. To effectively learn the change in angular differences, ACEV employs a neighbourhood-based approach using the Exponential Moving Average (EMA) method \cite{haynes2012exponential}. The EMA method is applied through the following equation \ref{EMA}. This strategic use of EMA ensures that the algorithm adapts to the evolving nature of local structures and captures the subtle variations in the angular gaps between eigenvectors. The predicted angular gap is computed as

\begin{equation}
\label{EMA}
\begin{split}
E_{d}(s) = \alpha \cdot {angle\_differ}_{d}(s, s-1) + \\
(1 - \alpha) \cdot E_{d}(s-1)
\end{split}
\end{equation}
where $E_{d}(s)$ predicts the angular difference of $d$th eigenvector for $s$th data point, $\alpha$ is the exponential smoothing factor and $d=1 \cdots D$.

To segment intersecting manifolds, ACEV initiates traversal from a data point $t \in l_i$, where $t = \min(t_1, \cdots, t_{n_i})$ for any one dimension in $D$ and $n_i$ is the number of unlabelled data points in $l_i$ which are not included in any individual manifold. $t$ is first considered for the manifold and serves as the root. After determining $t$, its $k$-nearest neighbor data points are identified for traversal and probable inclusion as its children. The algorithm aims to include data points within the same manifold that exhibit non-significant angular differences in all directions. It begins by finding the eigenvectors of the neighbourhood for both the parent $s-1$ and potential child data points $s$. Subsequently, it calculates the angular differences ${angle\_differ}_{g}(s, s-1)$ between these vectors using equation \ref{ang12}. ACEV then predicts angular differences $E_{d}(s)$ for all directions using equation \ref{EMA}. If the difference between $E_{d}(s)$ and ${angle\_differ}_{d}(s, s-1)$ is insignificant for all $D$ directions, then the potential child is included in the same manifold as the parent.

The inclusion process follows a depth-first search method, creating a tree structure with $t$ as the root. The calculation of equation \ref{EMA} follows the path from the root to the specific data point $s$ for which it is calculated and $s-1$ is the parent of $s$ and so on in the tree structure. This inclusion procedure continues until each data point in $l_i$ is included, or with the current neighbourhood, the probable child $s$ couldn't be included. 

It is important to mention that the EMA method requires an initial construction phase and therefore, $0.05\%$ of unlabelled data points are included in the tree without constraint. This step has negligible significance due to the inclusion of only $0.05\%$ of unlabelled data points and due to adaptation the effect also fades.

\subsubsection{Intersecting Neighbourhood Filtration}
The unsatisfiability of the inclusion criterion signifies that there are data points in the neighbourhood, for which there is a significant angular difference in one or more directions. This establishes that there is an increase in data variance in these directions, i.e., there is an increase in eigenvalue. It is evident that the probable child is lying in the intersection region and ACEV filters the neighbourhood and considers those data points lying on that particular manifold that it is currently traversing and excludes those data points from other manifolds.

% So, the ACEV finds the predicted eigenvalue using the equation \ref{EMA} like the predicted angular gap.

\begin{figure}
\includegraphics[width=0.9\linewidth]{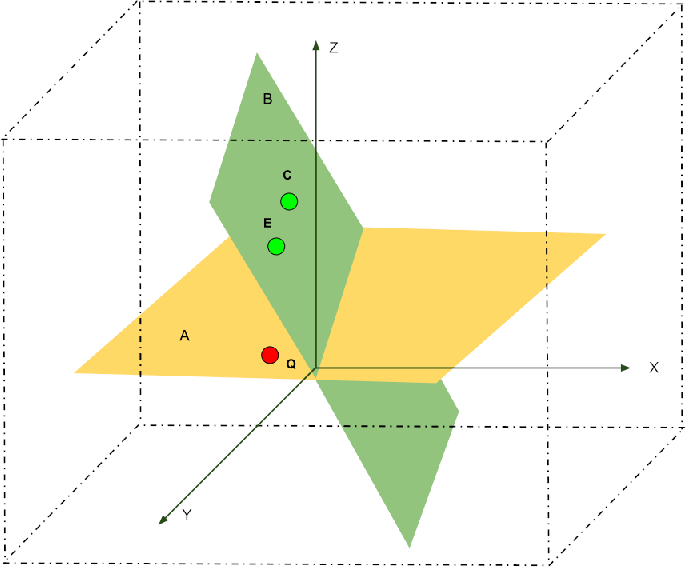} 
\caption{Eigenvalue and distance-based neighbourhood filtration}
\label{Fig 3}
\end{figure}

For example as shown in Figure \ref{Fig 3}, suppose $\Vec{X}$, $\Vec{Y}$ and $\Vec{Z}$ are the directions of data variance of $Q$, which lies on manifold $A$, and data points $C$ and $E$ are wrongly included in the neighbourhood of $Q$ and $Q$ couldn't be included in the manifold $A$. As $Q$ lies in manifold $A$, there will be zero eigenvalues in the direction of $Z$; therefore, the predicted angular gap in that direction will be nearly zero.

Now, consider the distances of $Q$, $C$ and $E$ from $\Vec{X}$, $\Vec{Y}$ and $\Vec{Z}$. It is clear that $Q$ will be very close to $\Vec{X}$ and $\Vec{Y}$ rather than $C$ and $E$, but may have similar distances from $\Vec{Z}$. In this scenario, for the removal of $C$ and $E$ from the neighbourhood of $Q$, eigenvector directions and associated eigenvalues of the covariance matrix of the neighbourhood of ($Q-1$), the parent of $Q$, will be beneficial. Let us consider the modified distance for each data point $r$ in the neighbourhood of $Q$ shown in the equation \ref{moddis}.
\begin{equation}
    mod\_dis(r)=\sum_{w=1}^{D} dis(e_w,r)\frac{E_{w}(r)}{E_{w}(Q-1)}.
    \label{moddis}
\end{equation}
where $dis(e_w,r)$ is the distance of datapoint $r$ from $w$th eigenvector of neighbourhood of ($Q-1$). $E_{w}(r)$ is eigenvalue for $w$th eigenvector of neighbourhood of $r$ and $E_{w}(Q-1)$ is eigenvalue for $w$th eigenvector of neighbourhood of ($Q-1$). The data points, which are on the same manifold, will have similar $E_{w}(r)$ with $E_{w}(Q-1), \forall w$ and $\frac{E_{w}(r)}{E_{w}(Q-1)}$ will be nearly $1$. In contrast, the data points which are not on the same manifold will have dissimilar values $E_{w}(r)$ with $E_{w}(Q-1)$ and $\frac{E_{w}(r)}{E_{w}(Q-1)}$ will be more than $1$. For example, in Figure \ref{Fig 3}, $C$ will have a higher variance in $\Vec{Z}$ direction, i.e., $E_{\Vec{Z}}(C)$ will be higher than $E_{\Vec{Z}}(Q-1)$ and $\frac{E_{\Vec{Z}}(C)}{E_{\Vec{Z}}(Q-1)}$ will be greater than $1$. This will be the contribution of $\frac{E_{w}(r)}{E_{w}(Q-1)}$ in demarcation. Similarly, the data points, which are not part of the manifold, may have similar $E_{w}(r)$ with $E_{w}(Q-1)$ but as they are not from the same manifold the value of $dis(e_w,r)$ will be higher than a data point which lies on that manifold. This will be the contribution of $dis(e_w,r)$ in demarcation. So, the $mod\_dis(r)$ for neighbourhood data points of $Q$ which are on manifold $A$ will be comparatively lower than data points $C$ and $E$. 

So using this intuition to filter the neighbourhood, $mod\_dis(r)$ is found for every data point in the neighbourhood of the probable child $Q$. Now, maintaining a decreasing order of $mod\_dis(r)$, the ACEV removes data points from the neighbourhood of $Q$ one by one and calculates the angular differences ${angle\_differ}_{d}(Q, Q-1), \forall d$ using equation \ref{ang12} with the updated neighbourhood. If the difference between $E_{d}(Q)$ and ${angle\_differ}_{d}(Q, Q-1)$ is insignificant for all directions $D$, the potential child $Q$ is included in the same manifold as the child. Removal of certain data points will satisfy the criterion, and $Q$ will be included with this updated neighbourhood and the traversal and inclusion of data points continue. This is how the ACEV finds individual manifolds and continues to search for manifolds in every non-intersecting component $l_i$ until all the data points become part of a manifold.

% \begin{figure}[h]
%          \fbox{\includegraphics[width=\linewidth, height=2in]{4.eps}}
%         %\caption{Caption for Image 2}
        
%         \caption{Eigenvalue and distance-based neighborhood filtration}
%         \label{Fig 3}
% \end{figure}

% \begin{figure*}[t]
%     \centering
%     \fbox{
%     \begin{minipage}{0.6\textwidth}
%         \centering
%         {\includegraphics[width=\linewidth]{4.eps}}
%     \end{minipage}%
%     \begin{minipage}{0.4\textwidth}
%         \caption{Eigenvalue and distance-based neighborhood filtration}
%         \label{Fig 3}
%     \end{minipage}
%     }
% \end{figure*}

\begin{figure*}[h]  % Correct way to specify figure placement
    \centering  % To center the figure
    \includegraphics[width=\textwidth, height = 3.5in]{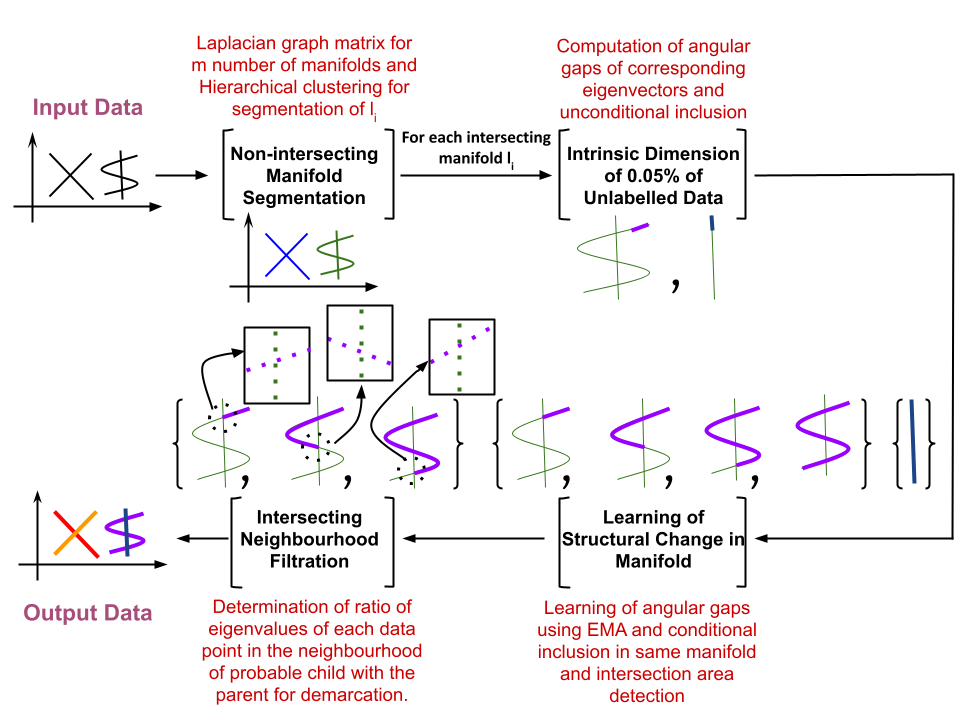}  % Correct specification of the image width
    \caption{Graphical workflow of the proposed method ACEV}
    \label{Fig 19}
\end{figure*}

Figure \ref{Fig 19} represents a concise workflow of ACEV. There are four intersecting manifolds, which belong to two non-intersecting manifolds. So, in the first step, two non-intersecting manifolds are separated. Next, considering one of them, the s-curve and line, ACEV starts segmenting intersecting ones. Initial $0.05\%$ of data points are labelled unconditionally and then, data points are included maintaining EMA-dependent angular gap and intersecting regions are detected. Upon detection, ACEV filters the neighbourhood, includes data points, and continues until no unlabelled data point can be included. Then, the next manifold is segmented without hindrance as the other intersecting manifold is labelled. For the other non-intersecting manifold, two individuals are segmented and ultimately, the four manifolds are segmented.
\subsection{Algorithm and Time Complexity Analysis}

The algorithm involves several steps: firstly, $O(nD \log(n))$ is needed to construct the tree-like structure of $k$-neighborhoods. Subsequently, $O(n^3)$ is required for labeling and segmenting non-intersecting manifolds. Finding the $k$ nearest neighbor of the root data point takes $O(k \log(n))$ time, followed by determining the principal axis, which consumes $O(n^3)$ time involving covariance matrix computation and eigenvector calculation. The angle between vectors is found in $O(D)$ time, where $D$ is the number of dimensions. Since, the algorithm runs recursively for each neighbor, the time required for each recursion is $\log_k{n}$. The overall time complexity of the ACEV is expressed as $O(nD \log(n) + n^3 + \log_k{n}(k \log(n) + n^3 + D))$. It's noteworthy that the complexity is expected to decrease over time as manifold determination reduces the number of data points, denoted as $n$.

\section{Performance Analysis and Comparative Study}
\subsection{Experimental Setup and Performance Evaluation Metrics}
The performance analysis and comparative study were carried out on an Intel $i5$ processor with a clock speed of $4.90$ GHz and $16$ gigabytes of RAM without a dedicated graphics processing unit. The effectiveness of the proposed method is examined using real-life datasets with diverse sample sizes, dimensions, and classes. Table \ref{tab:dataset} contains the description of each dataset. The exponential smoothing factor $\alpha$ was set to $0.6$ and the $k$-neighborhood value were employed for individual datasets with $k=25$.
  
  \begin{table}[h]
  \caption{DESCRIPTION OF THE BENCHMARK DATASETS}
  \centering
  \label{tab:dataset}
  \resizebox{\columnwidth}{!}{%
  \begin{tabular}{|c|c|c|c|c|} % Defines a table with four centered columns
    \hline
    \textbf{Sl No.} & \textbf{Dataset} & \textbf{Samples} & \textbf{Dimension} & \textbf{Classes} \\
    \hline
    1 & Ecoli  \cite{misc_ecoli_39}  & 336 & 7 & 8 \\
    \hline
    2 & Wine \cite{misc_wine_109} & 178 & 13 & 3 \\
    \hline
    3 & Car\cite{misc_car_evaluation_19} & 1728 & 6 & 4 \\
    \hline
    4 & Echocardiogram \cite{misc_echocardiogram_38} & 131 & 10 & 2 \\
    \hline
    5 & Ionosphere \cite{ionosphere_52}  & 351 & 33 & 2 \\
    \hline
    6 & Hepatitis \cite{misc_hepatitis_46} & 155 & 19 & 2 \\
    \hline
    7 & Zoo \cite{misc_zoo_111} & 101 & 16 & 7 \\
    \hline
    8 & Seeds \cite{charytanowicz2012seeds} & 210 & 7 & 3 \\
    \hline
    9 & Australia \cite{misc_statlog_(australian_credit_approval)_143} & 690 & 14 & 2 \\
    \hline
    10 & Iris \cite{misc_iris_53} & 150 & 4 & 3 \\
    \hline
    11 & Letter \cite{misc_letter_recognition_59} & 20000 & 16 & 12 \\
    \hline
    12 & DNA \cite{abcd} & 2000 & 180 & 3 \\
    \hline
    13 & Isolet \cite{misc_isolet_54} & 7797 & 617 & 26 \\
    \hline
    14 & Swarm Behaviour \cite{misc_swarm_behaviour_524} & 24017 & 2400 & 2 \\
    \hline
  \end{tabular}%
  }
\end{table}

% hardware setup + parameters value
% \subsection{Experimental Configuration and Parameterization}

% \subsection{Performance Evaluation Metrics}

For the evaluation of the performance of ACEV and the related state-of-the-art models, Adjusted Rand Index (ARI) \cite{santos2009use} and Normalized Mutual Information (NMI) \cite{koopman2017mutual} metrics have been used. These metrics assess the clustering efficiency by comparing the algorithm-determined clusters with the ground truth or reference clusters. 
The ARI considers the randomness in data point assignment to clusters and quantifies the clustering algorithm's ability to capture the true structure of the data. ARI ranges from $-1$ to $1$, where $1$ indicates a perfect match, $0$ suggests randomness and negative values imply clustering worse than random. NMI measures mutual information between true and predicted clustering. It normalizes the result between 0 and 1, considering the entropy of individual and joint clusterings. NMI ranges from $0$ to $1$, with higher values indicating better results. 

Two statistical analyses are considered to evaluate the qualitative development of ACEV. The Friedman test \cite{sheldon1996use} checks for performance differences among groups of ordinal data and calculates a Friedman statistic based on ranked data and the statistic is compared with the chi-squared distribution. If the calculated p-value is below the significance level ($0.05$) then the null hypothesis of no significant differences among groups is rejected and indicates dissimilarity. The Wilcoxon signed-rank test \cite{woolson2007wilcoxon} is used to ascertain whether ACEV's performance differs significantly from each of the other models if the Friedman test indicates that there are differences between the models. The Wilcoxon rank test compares paired observations to find significant differences. It involves the ranking of absolute differences between paired observations. The test statistic is calculated by summing the ranks of positive and negative differences separately. The p-value is then compared with the critical value $0.05$ to decide whether to reject the null hypothesis of no significant difference.

% The Friedman test is used to find whether the algorithms are the same or different. Friedman test is employed to find similarity or dissimilarity among the same groups of ordinal data. This test includes repeated measurements among each group calculating a Friedman statistic based on ranked data and comparing it with the chi-squared distribution. If the calculated p-value comes below a significance level, the null hypothesis of no significant differences among groups is rejected which clearly indicates the dissimilarity. On the other side non-rejection suggests similarity.
% Wilcoxon ranked test, which is a statistical method used to find whether there is a significant difference between paired observations. The test involves ranking the absolute differences between the paired observations, summing the rank of positive and negative differences separately, and calculating a test statistic. Then by comparing these test statistics, it is decided whether to reject the null hypothesis of no significant difference or not.

\subsection{Performance Analysis and Sensitivity Study on Parameters}

% The proposed method contains two parts, the first part contains the segmentation of non-intersecting manifolds and the second part contains the segmentation of the intersecting manifolds. As it is seen in Table \ref{tab:comparison}  both parts of the algorithm are required. For example, as it is seen in Table \ref{tab:comparison} for datasets 4, 7 and 8 non-intersecting manifold segmentation performs better than that of intersecting manifold segmentation. While in other datasets intersecting manifold segmentation performs better than non-intersecting manifold segmentation. But in all the cases the ACEV performs much better than either intersecting or non-intersecting. This clearly indicates that both intersecting and non-intersecting parts of the algorithm are required.
The proposed method consists of two components: the first focuses on segmenting non-intersecting manifolds, and the second on intersecting manifolds. Table \ref{tab:comparison} highlights the necessity of both parts, with non-intersecting manifold segmentation excelling for datasets Echocardiogram, Zoo, and Seeds, while intersecting manifold segmentation performs better for other datasets. Notably, ACEV consistently outperforms both individual approaches and highlights the necessity of incorporating both intersecting and non-intersecting manifold segmentation to achieve proper segmentation.

\begin{table}[h]
\centering
\caption{Comparative Analysis of Individual Segmentation Mechanism with ACEV}
\label{tab:comparison}
\resizebox{\columnwidth}{!}{%
\begin{tabular}{|c|cc|cc|cc|}
\hline
{\multirow{2}{*}{Sl No.}} & \multicolumn{2}{c|}{Non-Intersecting} & \multicolumn{2}{c|}{Intersecting} & \multicolumn{2}{c|}{\textbf{ACEV}} \\ \cline{2-7}
     & ARI                           & NMI                          & ARI                        & NMI                        & \multicolumn{1}{c|}{ARI}             & NMI            \\
\hline
1    & \multicolumn{1}{c|}{0.48}                         & 0.32                        & \multicolumn{1}{c|}{0.61}                       & 0.37                      & \multicolumn{1}{c|}{\textbf{0.81}}  & \textbf{0.68} \\\hline
2    & \multicolumn{1}{c|}{0.45}                        & 0.35                        & \multicolumn{1}{c|}{0.50}                   & 0.32                     & \multicolumn{1}{c|}{\textbf{0.80}}  & \textbf{0.69} \\\hline
3    & \multicolumn{1}{c|}{0.39}                         & 0.25                       & \multicolumn{1}{c|}{0.42}                        & 0.30                   & \multicolumn{1}{c|}{\textbf{0.68}}  & \textbf{0.50} \\\hline
4    & \multicolumn{1}{c|}{0.64}                         & 0.40                       & \multicolumn{1}{c|}{0.59}                    & 0.38                    & \multicolumn{1}{c|}{\textbf{0.89}}  & \textbf{0.63} \\\hline
5    & \multicolumn{1}{c|}{0.58}                        & 0.32                        & \multicolumn{1}{c|}{0.59}                     & 0.38                     & \multicolumn{1}{c|}{\textbf{0.72}}  & \textbf{0.50} \\\hline
6    & \multicolumn{1}{c|}{0.38}                        & 0.29                       & \multicolumn{1}{c|}{0.40}                    & 0.32                   & \multicolumn{1}{c|}{\textbf{0.64}}  & \textbf{0.43} \\\hline
7    & \multicolumn{1}{c|}{0.7}                          & 0.56                        & \multicolumn{1}{c|}{0.66}                      & 0.50                    & \multicolumn{1}{c|}{\textbf{0.9}}  & \textbf{0.89} \\\hline
8    & \multicolumn{1}{c|}{0.69}                        & 0.54                        & \multicolumn{1}{c|}{0.68}                     & 0.52                     & \multicolumn{1}{c|}{\textbf{0.87}}  & \textbf{0.863} \\\hline
9    & \multicolumn{1}{c|}{0.56}                        & 0.29                        & \multicolumn{1}{c|}{0.59}                    & 0.30                      & \multicolumn{1}{c|}{\textbf{0.7}}   & \textbf{0.54} \\\hline
10   & \multicolumn{1}{c|}{0.62}                        & 0.65                       & \multicolumn{1}{c|}{0.78}                     & 0.63                    & \multicolumn{1}{c|}{\textbf{0.89}}  & \textbf{0.9} \\\hline
11   & \multicolumn{1}{c|}{0.54}                         & 0.32                        & \multicolumn{1}{c|}{0.57}                    & 0.39                     & \multicolumn{1}{c|}{\textbf{0.78}}  & \textbf{0.63} \\\hline
12   & \multicolumn{1}{c|}{0.65}                        & 0.43                       & \multicolumn{1}{c|}{0.72}                     & 0.69                     & \multicolumn{1}{c|}{\textbf{0.97}}  & \textbf{0.93} \\\hline
13   & \multicolumn{1}{c|}{0.53}                         & 0.41                        & \multicolumn{1}{c|}{0.58}                      & 0.53                   & \multicolumn{1}{c|}{\textbf{0.75}}  & \textbf{0.77} \\\hline
14   & \multicolumn{1}{c|}{0.48}                        & 0.51                      & \multicolumn{1}{c|}{0.53}                      & 0.61                    & \multicolumn{1}{c|}{\textbf{0.84}}  & \textbf{0.86} \\\hline
\end{tabular}%
}
\end{table}

% \begin{figure}[h]

% \begin{subfigure}{0.5\textwidth}
% \includegraphics[width=\linewidth, height=1.5in]{learning1.png} 
% \caption{Performance dependency on training phase}
% \label{learn}
% \end{subfigure}
% \begin{subfigure}{0.5\textwidth}
% \includegraphics[width=\linewidth, height=1.5in]{alpha1.png}
% \caption{Performance dependency on exponential smoothing factor}
% \label{alpha}
% \end{subfigure}

% \caption{Sensitivity analysis of ACEV on two parameters}
% \label{fig:image2}
% \end{figure}

\begin{figure} 
    \centering
  \subfloat[Performance dependency on training phase\label{1a}]{%
       \includegraphics[width=\linewidth]{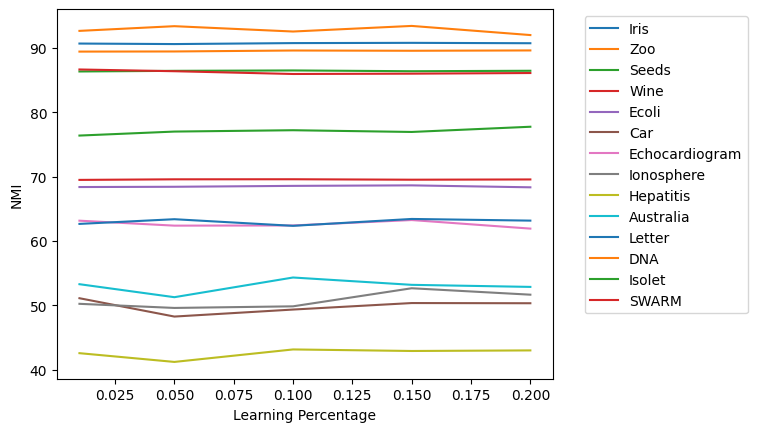}}
    \hfill
  \subfloat[Performance dependency on exponential smoothing factor\label{1b}]{%
        \includegraphics[width=\linewidth]{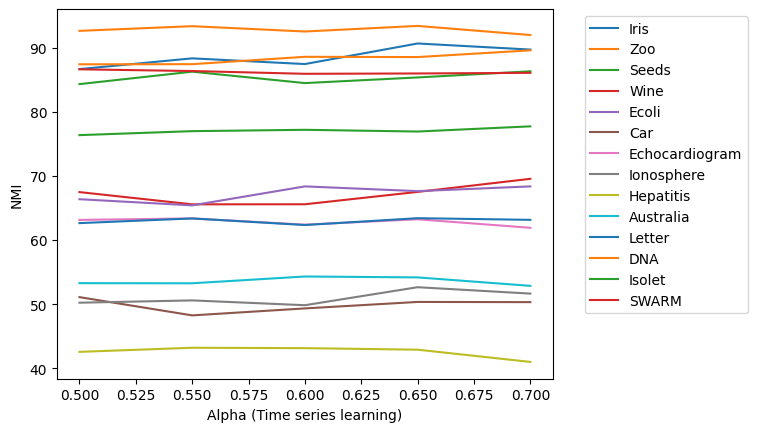}}
  \caption{Sensitivity analysis of ACEV on two parameters}
  \label{alpha} 
\end{figure}

\begin{figure}
\includegraphics[width=\linewidth]{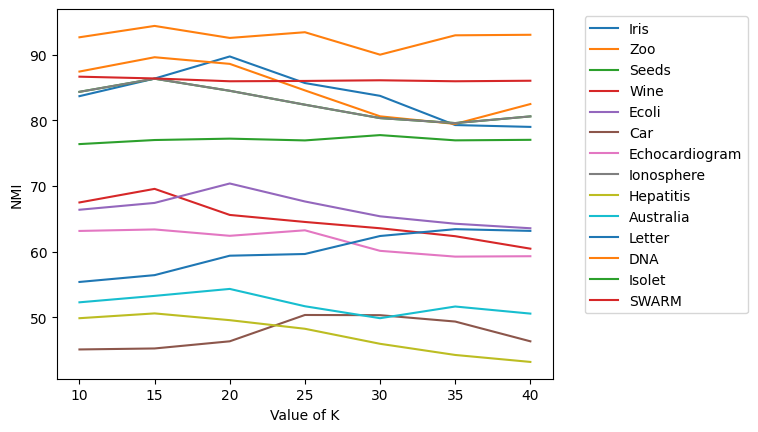} 
\caption{Performance dependency on nearest neighbor parameter}
\label{k}
\end{figure}

The efficacy of the ACEV depends on three key factors: $k$, $\alpha$, and the learning percentage. To assess its performance across these parameters, Figure \ref{alpha}, and \ref{k} illustrate the algorithm's behavior over a specified range for each variable. The algorithm indeed demonstrates robustness with minimal training requirements, as evidenced by the ablation study which indicates stability and independence across varying learning percentages. 
Notably, the algorithm's performance remains unaffected by the exponential smoothing factor and exhibits resilience to changes in the neighbourhood parameter. The latter, although recognized as an open research problem due to the delicate balance required for optimal performance, showcase the algorithm's effectiveness, as it operates reliably over a range of nearest-neighbor values. High values of $k$ are noted for potentially compromising the locally linear property of the manifold, while excessively low values risk generating disconnected components. Despite these challenges, the ACEV performs consistently and effectively, making it less vulnerable to fluctuations in the nearest neighbour parameter.
% \begin{figure}[h]

%     \centering
%     %\begin{subfigure}{0.45\textwidth}
%          \fbox{\includegraphics[width=\linewidth, height=2in]{learningpercentage.eps}}
%         \caption{Performance dependency on training phase}
%         \label{learn}
%     \end{figure}

\subsection{Comparative Analysis}
% ARI and NMI scores are found for various algorithms over different real-life datasets. The comparative analysis is given in Table: \ref{tab:performance1}, \ref{tab:performance2}, where the best performing algorithm is colored green for a specific dataset, and the second best performing algorithm is colored yellow. The analysis shows that for the majority of the datasets, the ACEV performs best and for which it performs second best remains very close to the best one.

% Then Friedman test with the ARI and NMI scores is performed. Friedman test with the ARI score of the Performance analysis table gave a P-value: $7.517839686041210e-16$ whereas Friedman test for NMI scores gave a P-value: $2.1601684883457277e-18$. It is clearly visible that Friedman test values for both the ARI and NMI are significantly less. Therefore all the clustering algorithms are different. In the last row of Tables \ref{tab:performance1} and \ref{tab:performance2}, the Wilcoxon rank test signifies how the proposed method is significantly different from others.

ARI and NMI scores have been computed for $18$ state-of-the-art intersecting manifold segmentation algorithms across real-life datasets. The performance of comparative methods and ACEV are presented in Tables \ref{tab:performance1} and \ref{tab:performance2}. In these tables, the best-performing algorithm is highlighted in green for each dataset, while the second-best is colored yellow. Notably, the ACEV consistently outperforms others, and even when it is the second-best, the performance is very close to the top algorithm.

Following this, a Friedman test was conducted using ARI and NMI scores. The obtained p-values for ARI and NMI were $7.52 \times 10^{-16}$ and $2.16 \times 10^{-18}$, respectively. These significantly low p-values indicate that clustering algorithms differ significantly. The last row of Tables \ref{tab:performance1} and \ref{tab:performance2} presents the results of the Wilcoxon rank test, emphasizing the significant difference of the proposed method from others. This signifies that the performance of ACEV is stable over these datasets whereas other methods sometimes outperform ACEV for a few datasets.

\begin{sidewaystable}[htbp]
    \centering
    \caption{Comparative Performance Analysis of State-of-the-art Methods and ACEV Using Adjusted Rand Index (ARI) Scores}
    \label{tab:performance1}
    \resizebox{\textwidth}{!}{
        \begin{tabular}{|c|c|c|c|c|c|c|c|c|c|c|c|c|c|c|c|c|c|c|c|}
            \hline
             SL No. & $k$means  & Mean Shift  & Optics  & Birch  & DBSCAN  & SMMC  & KPC  & KPPC  & LKPPC  & TWSVC  & KFC  & LKFC  & RTWSVC  & KSDC  & UMD Isomap  & Graph MMC  & TVG+AR & MFPC  & \textbf{ACEV} \\
            \hline
            1 & 0.70 & 0.63 & 0.41 & 0.50 & 0.46 & 0.50 & 0.51 & 0.51 &\textbf{\textcolor[RGB]{255, 191, 0}{0.84}} & 0.82 & 0.70 & \textbf{\textcolor[RGB]{255, 191, 0}{0.84}} & 0.52 & 0.77 & 0.65 & 0.72 & 0.56 & \textbf{\textcolor[RGB]{0, 173, 96}{0.86}} & 0.81 \\
            \hline
            2 & 0.68 & 0.51 & 0.39 & 0.45 & 0.50 & 0.69 & 0.51 & 0.52 & 0.71 & 0.67 & 0.67 & 0.68 & 0.68 & 0.66 & 0.74 & 0.68 & 0.49 &\textbf{\textcolor[RGB]{255, 191, 0}{0.77}} & \textbf{\textcolor[RGB]{0, 173, 96}{0.80}} \\
            \hline
            3 & 0.54 & 0.50 & 0.41 & 0.49 & 0.45 & 0.51 & 0.52 & 0.56 & 0.58 & 0.53 & 0.54 & 0.60 & 0.52 & 0.54 & 0.55 & 0.56 & 0.36 &\textbf{\textcolor[RGB]{255, 191, 0}{0.61}} & \textbf{\textcolor[RGB]{0, 173, 96}{0.68}} \\
            \hline
            4 & 0.68 & 0.61 & 0.50 & 0.56 & 0.57 & 0.76 & 0.51 & 0.54 & 0.73 & 0.50 & 0.72 & 0.72 & 0.73 & 0.68 & 0.76 & 0.70 & 0.39 &\textbf{\textcolor[RGB]{255, 191, 0}{0.86}} & \textbf{\textcolor[RGB]{0, 173, 96}{0.89}} \\
            \hline
            5 & 0.57 & 0.50 & 0.36 & 0.48 & 0.40 & 0.67 & 0.61 & 0.53 & 0.59 & 0.50 &\textbf{\textcolor[RGB]{255, 191, 0}{0.70}} & 0.60 & 0.66 & 0.57 & 0.53 & 0.56 & 0.35 & 0.59 & \textbf{\textcolor[RGB]{0, 173, 96}{0.72}} \\
            \hline
            6 & 0.51 & 0.53 & 0.45 & 0.51 & 0.47 & 0.49 & 0.47 & 0.52 & 0.50 & 0.50 & 0.52 & 0.50 & \textbf{\textcolor[RGB]{255, 191, 0}{0.63}} & 0.50 & 0.49 & 0.52 & 0.50 &{0.52} & \textbf{\textcolor[RGB]{0, 173, 96}{0.64}} \\
            \hline
            7 & 0.81 & 0.73 & 0.60 & 0.59 & 0.62 & 0.78 & 0.61 & 0.75 & {0.85} & 0.83 & 0.85 & 0.91 & 0.61 & 0.84 & 0.73 & 0.73 & 0.44 & \textbf{\textcolor[RGB]{0, 173, 96}{0.96}} &\textbf{\textcolor[RGB]{255, 191, 0}{0.90}} \\
            \hline
            8 & 0.85 & 0.70 & 0.55 & 0.60 & 0.45 & 0.81 & 0.72 & 0.60 & 0.84 & 0.65 & 0.72 & 0.85 & 0.72 & 0.85 & 0.69 & 0.72 & 0.53 & \textbf{\textcolor[RGB]{0, 173, 96}{0.94}} &\textbf{\textcolor[RGB]{255, 191, 0}{0.86}} \\
            \hline
            9 & 0.50 & 0.45 & 0.29 & 0.35 & 0.30 & 0.50 & 0.49 & 0.50 & 0.50 & 0.50 & 0.52 & 0.50 & 0.49 & 0.50 & 0.60 & \textbf{\textcolor[RGB]{255, 191, 0}{0.62}} & 0.44 &{0.61} & \textbf{\textcolor[RGB]{0, 173, 96}{0.70}} \\
            \hline
            10 & 0.86 & 0.79 & 0.60 & 0.74 & 0.45 & 0.85 & 0.63 & 0.56 & \textbf{\textcolor[RGB]{255, 191, 0}{0.95}} & 0.90 & 0.90 & 0.87 &{0.91} & 0.89 & 0.74 & 0.76 & 0.62 & \textbf{\textcolor[RGB]{0, 173, 96}{0.98}} & 0.89 \\
            \hline
            11 & 0.68 & 0.63 & 0.51 & 0.60 & 0.55 & 0.53 & 0.50 & 0.51 & 0.74 & 0.56 & 0.62 & 0.68 & 0.60 & 0.63 & 0.63 & 0.65 & 0.62 &\textbf{\textcolor[RGB]{255, 191, 0}{0.76}} & \textbf{\textcolor[RGB]{0, 173, 96}{0.78}} \\
            \hline
            12 & 0.63 & 0.60 & 0.46 & 0.54 & 0.50 & 0.75 & 0.75 & 0.69 &\textbf{\textcolor[RGB]{255, 191, 0}{0.96}} & 0.74 & 0.76 &\textbf{\textcolor[RGB]{255, 191, 0}{0.96}} & 0.74 &\textbf{\textcolor[RGB]{255, 191, 0}{0.96}} & 0.81 & 0.83 & 0.57 &\textbf{\textcolor[RGB]{255, 191, 0}{0.96}} & \textbf{\textcolor[RGB]{0, 173, 96}{0.97}} \\
            \hline
            13 & 0.53 & 0.50 & 0.45 & 0.49 & 0.42 & \textbf{\textcolor[RGB]{255, 191, 0}{0.80}} & 0.70 & 0.59 & 0.68 & 0.63 & 0.65 & 0.63 & 0.66 & .59 & 0.58 & 0.60 & 0.45 &{0.67} & \textbf{\textcolor[RGB]{0, 173, 96}{0.85}} \\
            \hline
            14 & 0.60 & 0.54 & 0.50 & 0.52 & 0.54 & 0.61 & 0.49 & 0.53 & 0.73 & 0.58 & 0.74 & 0.73 & 0.69 & 0.78 & 0.70 & 0.71 & 0.40 &\textbf{\textcolor[RGB]{255, 191, 0}{0.80}} & \textbf{\textcolor[RGB]{0, 173, 96}{0.84}} \\
            \hline
            P-Value & .000122 & 0.00012207 & 0.00012207 & 0.00012207 & 0.00012207 & 0.000366 & 0.000122 & 0.000122 & 0.00671387 & 0.000610352 & 0.00024414 & 0.003051758 & 0.000366211 & 0.00024414 & 0.00012204 & 0.000122 & 0.00012204 & 0.02121094 & \\
            \hline
        \end{tabular}%
    }
% \end{table*}
\end{sidewaystable}

% Please add the following required packages to your document preamble:
% \usepackage[table,xcdraw]{xcolor}
% Beamer presentation requires \usepackage{colortbl} instead of \usepackage[table,xcdraw]{xcolor}
\begin{sidewaystable}[htbp]
    \centering
    \caption{Comparative Performance Analysis of State-of-the-art Methods and ACEV Using Normalized Mutual Information (NMI) Scores}
    \label{tab:performance2}
    \resizebox{\textwidth}{!}{%
    \begin{tabular}{|c|c|c|c|c|c|c|c|c|c|c|c|c|c|c|c|c|c|c|c|}
        \hline
        SL No. & $k$means  & Mean Shift  & Optics  & Birch  & DBSCAN  & SMMC  & KPC  & KPPC  & LKPPC  & TWSVC  & KFC  & LKFC  & RTWSVC & KSDC  & UMD Isomap  & Graph MMC  & TVG+AR  & MFPC  & \textbf{ACEV} \\
        \hline
        1 & 0.59 & 0.52 & 0.48 & 0.50 & 0.36 & 0 & 0.16 & 0.21 & 0.66 & 0.58 & 0.52 & 0.65 & 0.20 & 0.61 & 0.60 & 0.51 & 0.43 & \textbf{\textcolor[RGB]{255, 191, 0}{0.68}} & \textbf{\textcolor[RGB]{0, 173, 96}{0.69}} \\
        \hline
        2 & 0.42 & 0.38 & 0.35 & 0.40 & 0.34 & 0.39 & 0.08 & 0.52 & 0.47 & 0.49 & 0.43 & 0.44 & 0.45 & 0.40 & 0.58 & 0.53 & 0.35 & \textbf{\textcolor[RGB]{255, 191, 0}{0.67}} & \textbf{\textcolor[RGB]{0, 173, 96}{0.69}} \\
        \hline
        3 & 0.33 & 0.20 & 0.15 & 0.18 & 0.10 & 0.11 & 0.08 & 0.19 & 0.31 & 0.18 & 0.14 & 0.28 & 0.16 & 0.19 & 0.39 & \textbf{\textcolor[RGB]{255, 191, 0}{0.44}} & 0.34 & {0.29} & \textbf{\textcolor[RGB]{0, 173, 96}{0.50}} \\
        \hline
        4 & 0.32 & 0.30 & 0.30 & 0.25 & 0.23 & 0.48 & 0.58 & 0.03 & 0.41 & 0.09 & 0.39 & 0.39 & 0.45 & 0.37 & 0.52 & 0.51 & 0.44 & \textbf{\textcolor[RGB]{0, 173, 96}{0.67}} & \textbf{\textcolor[RGB]{255, 191, 0}{0.63}} \\
        \hline
        5 & 0.12 & 0.25 & 0.22 & 0.20 & 0.21 & 0.27 & 0.14 & 0.03 & 0.13 & 0.02 & {0.31} & 0.26 & 0.26 & 0.11 & 0.38 & \textbf{\textcolor[RGB]{255, 191, 0}{0.46}} & 0.25 & 0.13 & \textbf{\textcolor[RGB]{0, 173, 96}{0.50}} \\
        \hline
        6 & 0.003 & 0.02 & 0 & 0.05 & 0 & 0 & 0.01 & 0.09 & 0.003 & 0.003 & 0.008 & 0.004 & 0.15 & 0.003 & 0.28 & \textbf{\textcolor[RGB]{255, 191, 0}{0.37}} & 0.29 & {0.07} & \textbf{\textcolor[RGB]{0, 173, 96}{0.43}} \\
        \hline
        7 & 0.73 & 0.60 & 0.49 & 0.52 & 0.50 & 0.73 & 0.50 & 0.57 & 0.78 & 0.74 & 0.80 & {0.82} & 0.50 & 0.77 & 0.71 & 0.61 & 0.64 & \textbf{\textcolor[RGB]{255, 191, 0}{0.893}} & \textbf{\textcolor[RGB]{0, 173, 96}{0.898}} \\
        \hline
        8 & 0.70 & 0.59 & 0.55 & 0.50 & 0.45 & 0.64 & 0.51 & 0.20 & 0.72 & 0.42 & 0.52 & 0.69 & 0.52 & 0.66 & 0.69 & 0.69 & 0.56 & \textbf{\textcolor[RGB]{255, 191, 0}{0.84}} & \textbf{\textcolor[RGB]{0, 173, 96}{0.86}} \\
        \hline
        9 & 0.03 & 0.10 & 0.05 & 0.02 & 0.07 & 0.03 & 0.32 & 0.01 & 0.02 & 0.02 & 0.02 & 0.02 & 0.008 & 0.02 & 0.04 & \textbf{\textcolor[RGB]{255, 191, 0}{0.49}} & 0.42 & {0.24} & \textbf{\textcolor[RGB]{0, 173, 96}{0.54}} \\
        \hline
        10 & 0.75 & 0.65 & 0.60 & 0.66 & 0.62 & 0.76 & 0.25 & 0.13 & 0.88 & 0.83 & 0.80 & 0.77 & 0.82 & 0.77 & 0.68 & 0.73 & 0.66 & \textbf{\textcolor[RGB]{0, 173, 96}{0.94}} & \textbf{\textcolor[RGB]{255, 191, 0}{0.90}} \\
        \hline
        11 & 0.42 & 0.40 & 0.35 & 0.41 & 0.39 & 0.20 & 0.01 & 0.05 & 0.54 & 0.25 & 0.34 & 0.42 & 0.43 & 0.33 & 0.54 & \textbf{\textcolor[RGB]{255, 191, 0}{0.56}} & 0.37 & \textbf{\textcolor[RGB]{255, 191, 0}{0.56}} & \textbf{\textcolor[RGB]{0, 173, 96}{0.63}} \\
        \hline
        12 & 0.36 & 0.45 & 0.40 & 0.47 & 0.42 & 0.55 & 0.58 & 0.42 & \textbf{\textcolor[RGB]{255, 191, 0}{0.91}} & 0.58 & 0.70 & \textbf{\textcolor[RGB]{255, 191, 0}{0.91}} & 0.58 & \textbf{\textcolor[RGB]{255, 191, 0}{0.91}} & 0.72 & 0.74 & 0.59 & \textbf{\textcolor[RGB]{255, 191, 0}{0.91}} & \textbf{\textcolor[RGB]{0, 173, 96}{0.93}} \\
        \hline
        13 & 0.40 & 0.31 & 0.35 & 0.32 & 0.29 & 0.48 & 0.45 & 0.51 & 0.62 & 0.55 & 0.62 & 0.50 & 0.61 & 0.52 & 0.59 & \textbf{\textcolor[RGB]{255, 191, 0}{0.70}} & 0.42 & {0.64} & \textbf{\textcolor[RGB]{0, 173, 96}{0.77}} \\
        \hline
        14 & 0.52 & 0.49 & 0.40 & 0.42 & 0.35 & 0.61 & 0.59 & 0.39 & 0.71 & 0.40 & 0.66 & 0.65 & 0.66 & 0.75 & 0.73 & 0.77 & 0.55 & \textbf{\textcolor[RGB]{255, 191, 0}{0.82}} & \textbf{\textcolor[RGB]{0, 173, 96}{0.86}} \\
        \hline
        P-Value & 0.000122 & 0.00012207 & 0.00012207 & 0.00012207 & 0.00012207 & 0.000122 & 0.000122 & 0.000122 & 0.00012207 & 0.00012207 & 0.00012207 & 0.00012207 & 0.00012207 & 0.00012207 & 0.012206 & 0.00012207 & 0.000122 & 0.01342773 & \\
        \hline
    \end{tabular}%
    }
\end{sidewaystable}

\section{Discussion}
% The proposed intersecting manifold segmentation mechanism incorporates two step segmentation; one is non-intersecting manifold segmentation and intersecting manifold segmentation. The performance analysis shows the requirements of both the steps. It also shows the performance efficiency over existing works and the time complexity of the proposed mechanism is better. The unsupervised segmentation ability of the proposed work makes the algorithm effective for real life incorporation.

% \begin{figure}[h]

%     \centering
%     %\begin{subfigure}{0.45\textwidth}
%         \includegraphics[width=\linewidth, height=.6in]{EMA2.eps}
%         \caption{Dependency on neighborhood construction for labeling co-existing concave and convex surface in same manifold}
%         \label{learn}
%     \end{figure}

\begin{figure}
\includegraphics[width=0.9\linewidth]{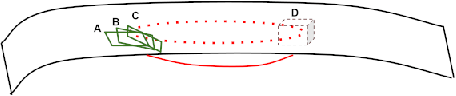} 
\caption{Dependency on neighbourhood construction for labelling co-existing concave and convex surface in same manifold}
\label{limit}
\end{figure}
An implicit assumption of the algorithm is that change in the tangent space should be smooth which may not be always satisfied. For example, if there is a concave part on a generally convex surface, then there is an abrupt change in the tangent direction as shown in Figure \ref{limit}. The proposed algorithm using EMA will learn the structural change in the tangent space shown in red in Figure \ref{limit} by learning the local structures $A$, $B$, and $C$ gradually and include them in the same manifold. This will depend on the neighbourhood construction because if it creates a neighbourhood like $D$ then there will be a significant difference in the tangent space and it will not be labelled as the same manifold. So the dependency of the algorithm like other existing algorithms on the neighborhood construction is a limitation.

\section{Conclusion and Future Work}

The proposed two-step intersecting manifold segmentation mechanism (ACEV) learns the intrinsic dimension of individual manifolds and segments them from each other manifolds and demonstrates notable efficiency gain over existing methods with better time complexity. The unsupervised segmentation capability makes ACEV well-suited for practical applications in real-life scenarios. Along with these positivities, the limitation will be reduced for betterment in the future.

% \section{Acknowledgment}
% This work was supported by the Technology Innovation Hub at ISI Kolkata on Data Science, Big Data Analytics, and Data Curation under Grant NMICPS/006/MD/2020-21 dt 16.10.2020 of DST, India.

\bibliographystyle{IEEEtran}
\bibliography{main}
\begin{IEEEbiography} [{\includegraphics[width=1in,height=1.25in,clip,keepaspectratio]{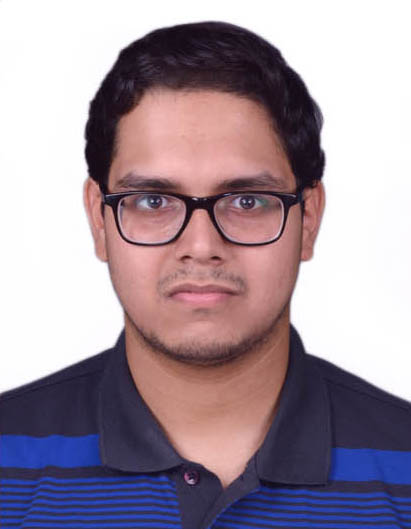}}] 
 {Subhadip Boral} received the M. Sc. in Computer Science degree from West Bengal State University in 2016. He is presently working as a Research Fellow at the Technology Innovation Hub at Indian Statistical Institute. His current research interests include Unsupervised Learning, Dimensionality Reduction, Streaming Data Analysis, and Anomaly Detection.
 \end{IEEEbiography}
\begin{IEEEbiography} [{\includegraphics[width=1in,height=1.25in,clip,keepaspectratio]{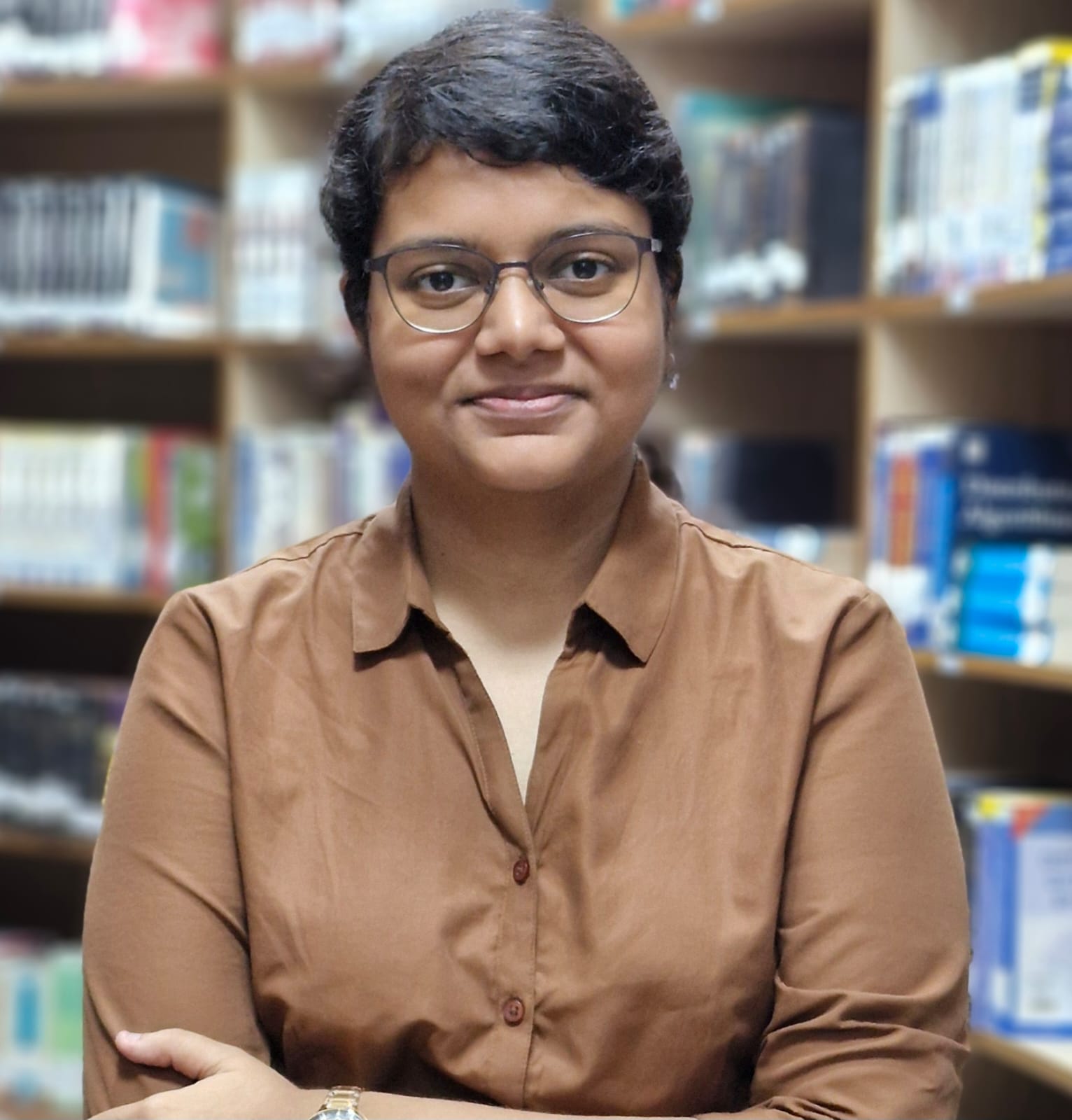}}] 
 {Rikathi Pal} completed her B.Tech in Information Technology from the University of Calcutta, Kolkata. She is currently a Pre-doctoral Research Fellow at the Indian Institute of Science (IISc), Bangalore. Her research focuses on image segmentation and the theoretical understanding of neural network training processes through topological data analysis (TDA). 
 \end{IEEEbiography}
 %Combining her expertise in TDA with neural networks, she aims to uncover deep insights into model optimization, contributing to advancements in both image processing and machine learning.

\begin{IEEEbiography} [{\includegraphics[width=1in,height=1.25in,clip,keepaspectratio]{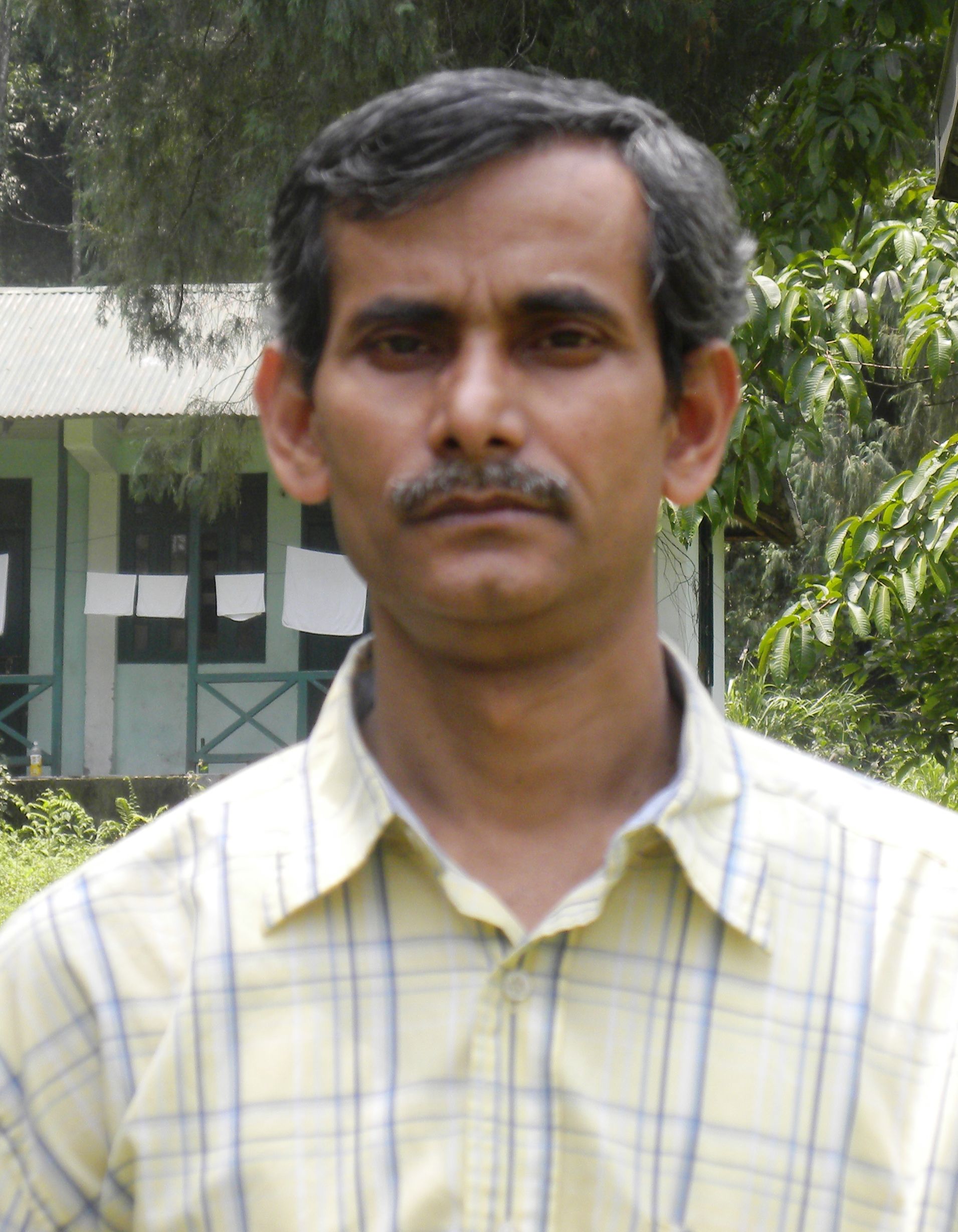}}] 
 {Ashish Ghosh} is a Senior Professor at the Indian Statistical Institute. He has published more than 270 research papers in international journals and conferences, and acting as the Principal Investigator of several funded projects. His current research interests include  Machine and Deep Learning, Data Science, Image/Video Analysis, and Computational Intelligence.
 \end{IEEEbiography}

% \bibliographystyle{elsarticle-num}
% \bibliography{main}
% \par\noindent 
% \parbox[t]{\linewidth}{
% \noindent\parpic{\includegraphics[height=1.5in,width=1in,clip,keepaspectratio]{Subhadip.jpg}}
% \noindent {\bf Subhadip Boral}\
% received the M. Sc. in Computer Science degree from West Bengal State University in 2016. He is presently working as a Research Fellow at the Technology Innovation Hub in Indian Statistical Institute. His current research interest includes Unsupervised Learning, Dimensionality Reduction, Streaming Data Analysis and Anomaly Detection}
% \vspace{1\baselineskip}
% \par\noindent 
% \parbox[t]{\linewidth}{
% \noindent\parpic{\includegraphics[height=1.5in,width=1in,clip,keepaspectratio]{Kousik.jpeg}}
% \noindent {\bf Kousik Roy}\
% received the B.E in Electrical Engineering degree from Jadavpur University, Kolkata, West Bengal in 2021. He is currently working as a Data Analyst. His current research interest includes Machine learning, deep learning and Time series analysis.}
% \vspace{1\baselineskip}
% \par\noindent 
% \parbox[t]{\linewidth}{
% \noindent\parpic{\includegraphics[height=1.5in,width=1in,clip,keepaspectratio]{A GHOSH 223.jpg}}
% \noindent {\bf Ashish Ghosh}\
% is a Senior Professor at the Indian Statistical Institute. He has published more than 270 research papers in international journals and conferences, and acting as the Principal Investigator of several funded projects. His current research interests include  Machine and Deep Learning, Data Science, Image/Video Analysis, and Computational Intelligence.}
%\vspace{4\baselineskip}
\end{document}